\documentclass[10pt,twocolumn,letterpaper]{article}

\usepackage{cvpr}              
\usepackage{graphicx}
\usepackage{amsmath}
\usepackage{amssymb}
\usepackage{booktabs}
\usepackage{enumitem}
\usepackage[accsupp]{axessibility}

\usepackage[pagebackref,breaklinks,colorlinks]{hyperref}

\usepackage[capitalize]{cleveref}
\crefname{section}{Sec.}{Secs.}
\Crefname{section}{Section}{Sections}
\Crefname{table}{Table}{Tables}
\crefname{table}{Tab.}{Tabs.}

\def\cvprPaperID{8301} 
\def\confName{CVPR}
\def\confYear{2023}
\def\Methodname{STSSL}

\begin{document}

\title{Spatiotemporal Self-supervised Learning for Point Clouds in the Wild}

\author{
Yanhao Wu $^{1}$ ~
Tong Zhang$^{2}$ ~
Wei Ke$^{1}$ ~
Sabine Süsstrunk$^{2}$ ~
Mathieu Salzmann$^{2}$\\
$^1$ School of Software Engineering, Xi'an Jiaotong University, China \\
$^2$ School of Computer and Communication Sciences, EPFL Switzerland \\
}

\maketitle

\begin{abstract}
Self-supervised learning (SSL) has the potential to benefit many applications, particularly those where manually annotating data is cumbersome. One such situation is the semantic segmentation of point clouds. In this context, existing methods employ contrastive learning strategies and define positive pairs by performing various augmentation of point clusters in a single frame. As such, these methods do not exploit the temporal nature of LiDAR data. In this paper, we introduce an SSL strategy that leverages positive pairs in both the spatial and temporal domain. To this end, we design (i) a point-to-cluster learning strategy that aggregates spatial information to distinguish objects; and (ii) a cluster-to-cluster learning strategy based on unsupervised object tracking that exploits temporal correspondences. We demonstrate the benefits of our approach via extensive experiments performed by self-supervised training on two large-scale LiDAR datasets and transferring the resulting models to other point cloud segmentation benchmarks. Our results evidence that our method outperforms the state-of-the-art point cloud SSL methods.~\footnote{Our code and pretrained models will be found at \href{https://github.com/YanhaoWu/STSSL}{https://github.com/YanhaoWu/STSSL}. Correspondence to Ke Wei.}



\end{abstract}

\vspace{-0.75em}

\section{Introduction}
\vspace{-0.5em}

\label{sec:intro}
Semantic segmentation from LiDAR point clouds can be highly beneficial in practical applications, e.g., for self-driving vehicles to safely interact with their surroundings. Nowadays, state-of-the-art methods~\cite{segmentation1, segmentation2, segmentation3} achieve this with deep neural networks. While effective, the training of such semantic segmentation networks requires large amounts of annotated data, which is prohibitively costly to acquire, particularly for point-level LiDAR annotations~\cite{depthcontrast}. By contrast, with the rapid proliferation of self-driving vehicles, large amounts of \emph{unlabeled} LiDAR data are generated. Here, we develop a method to exploit such unlabeled data in a self-supervised learning framework.

\begin{figure}[t]
  \centering
  \includegraphics[width=1\linewidth]{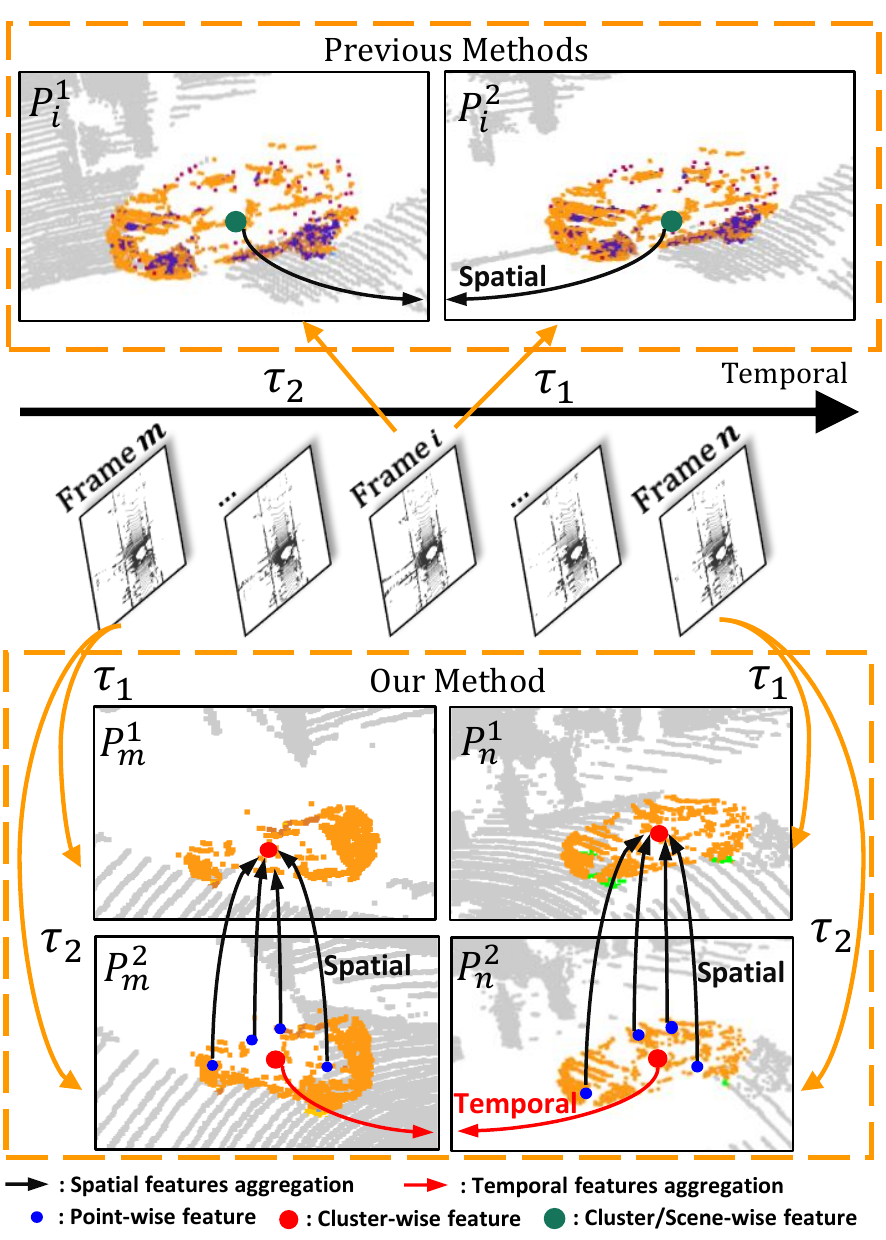}
  \caption{{\bf Our method vs existing ones.} {\bf (Top)} Previous methods create positive pairs for SSL by applying different augmentations, $\tau_{1}$ and $\tau_{2}$ (e.g., random flipping, clipping), to a single frame. {\bf (Bottom)} By contrast, we leverage both spatial and temporal information via a point-to-cluster and an inter-frame SSL strategy. Points in the same color are from the same cluster in the latent space.  
  }
  \vspace{-0.6cm}

  \label{fig:overview}
\end{figure}

\begin{figure}[t]
  \centering
  \includegraphics[width=1\linewidth]{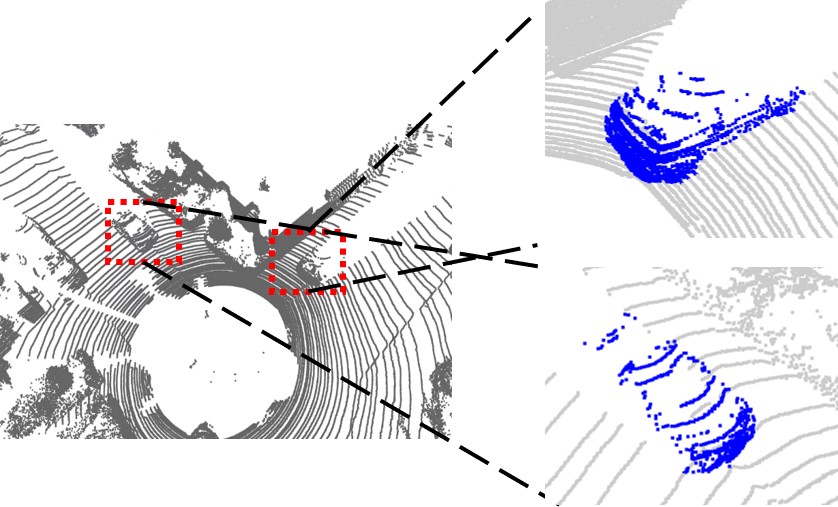}
  \caption{Cars in the {\bf same frame but under different illumination angles}. Note that the main source of difference between the two instance point clouds arises from the different illumination angles.}
  \label{fig:angle_different_1}
  \vspace{-0.6cm}

\end{figure}

Self-supervised learning (SSL) aims to learn features without any human annotations~\cite{depthcontrast, pointcontrast, segcontrast, pointssl1, pointssl2, pointssl3, pointssl4, pointssl5} but so that they can be effectively used for fine-tuning on a downstream task with a small number of labeled samples. This is achieved by defining a pre-task that does not require annotations. While many pre-tasks have been proposed~\cite{ssl_survey}, contrastive learning has nowadays become a highly popular choice~\cite{depthcontrast, gcc3d, proposalcontrast, segcontrast, pointcontrast}. In general, it aims to maximize the similarity of positive pairs while potentially minimizing that of negative ones. In this context, most of the point cloud SSL literature focuses on indoor scenes, for which relatively dense point clouds are available. Unfortunately, for outdoor scenes, such as the ones we consider here, the data is more complex and much sparser, and creating effective pairs remains a challenge.



Several approaches~\cite{segcontrast,depthcontrast} have nonetheless been proposed to perform SSL on outdoor LiDAR point cloud data.  As illustrated in the top portion of Fig.~\ref{fig:overview}, they construct positive pairs of point clusters or scenes by applying augmentations to a single frame. As such, they neglect the temporal information of the LiDAR data.
By contrast, in this paper, we introduce an SSL approach to LiDAR point cloud segmentation based on extracting effective positive pairs in both the spatial \emph{and temporal} domain. 

To achieve this without requiring any pose sensor as in~\cite{pointcontrast,STRL}, we introduce (i) a {\bf point-to-cluster (P2C)} SSL strategy that maximizes the similarity between the features encoding a cluster and those of its individual points, thus encouraging the points belonging to the same object to be close in feature space; (ii) a {\bf cluster-level inter-frame self-supervised learning} strategy that tracks an object across consecutive frames in an unsupervised manner and encourages feature similarity between the different frames. These two strategies are depicted in the bottom portion of Fig.~\ref{fig:overview}. 

Note that the illumination angle of one object seen in two different frames typically differs. As shown in Fig.~\ref{fig:angle_different_1}, this is also the main source of difference between two objects of the same class in the same frame. Therefore, our inter-frame SSL strategy lets us encode not only temporal information, but also the fact that points from different objects from the same class should be close to each other in feature space. As simulating different illumination angles via data augmentation is challenging, our approach yields positive pairs that better reflects the intra-class variations in LiDAR point clouds than existing single-frame methods~\cite{segcontrast,depthcontrast}.

Our contribution can be summarized as follows:
\begin{itemize}
\vspace{-0.5em}
\item We introduce an SSL strategy for point cloud segmentation based only on positive pairs. It does not require any external information, such as pose, GPS, and IMU.\vspace{-0.5em}
\item We propose a novel Point-to-Cluster (P2C) training paradigm that combines the advantages of point-level and cluster-level representations to learn a structured point-level embedding space.
\vspace{-0.5em}
\item We introduce the use of cluster-level inter-frame self-supervised leaning on point clouds generated by a LiDAR sensor, which introduces a new way to integrate temporal information into SSL.
\vspace{-0.5em}
\end{itemize}
Our experiments on several datasets, including KITTI\cite{kitti}, nuScene\cite{nuscene}, SemanticKITTI\cite{semantickitti} and SemanticPOSS\cite{semanticposs}, evidence that our method outperforms the state-of-the-art SSL techniques for point cloud data.

\begin{figure*}[t]
  \centering
  \includegraphics[width=1\linewidth]{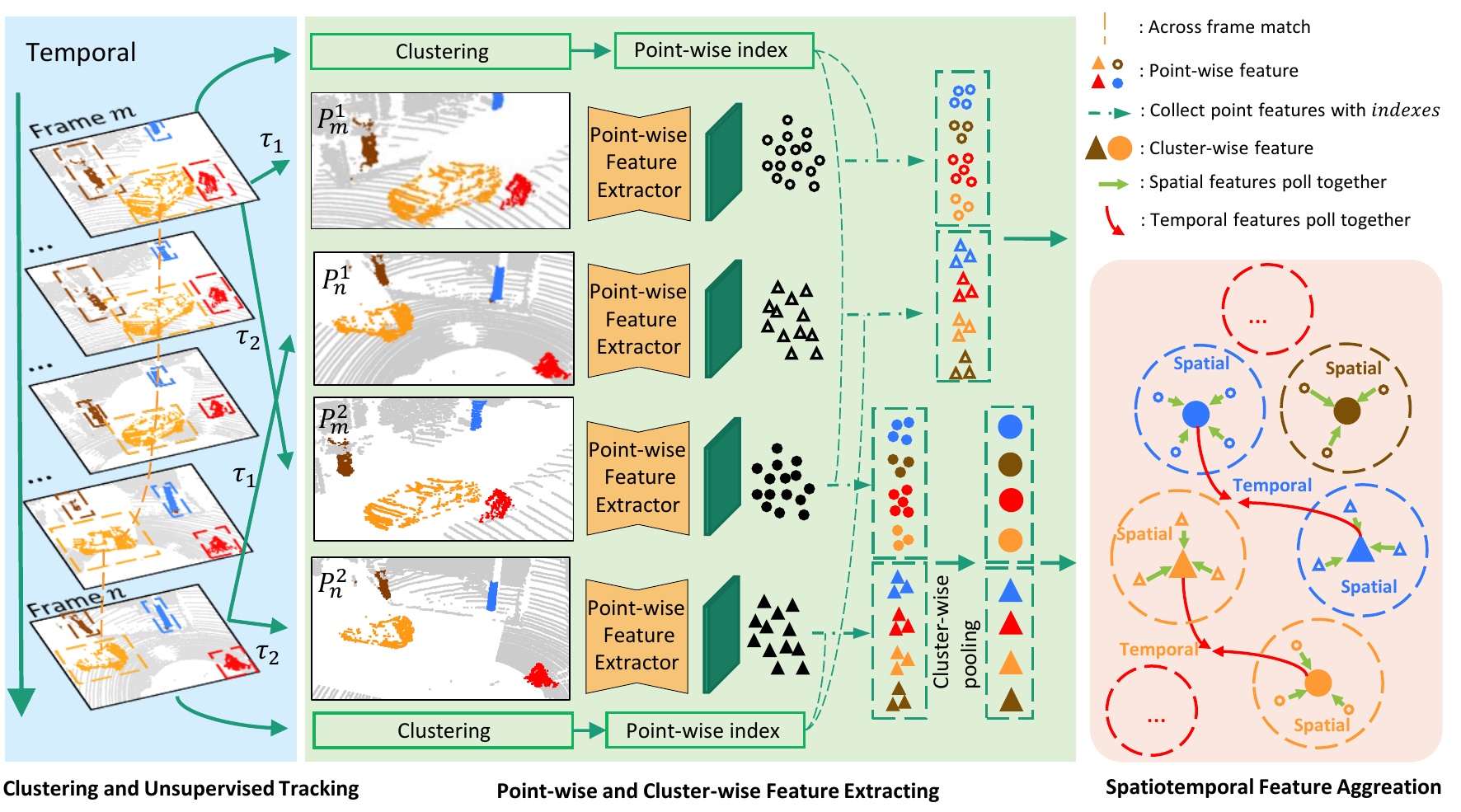}
  \vspace{-0.6cm}

  \caption{{\bf Overview of our \Methodname.} Given a sequence of LiDAR point clouds, we first perform clustering and unsupervised tracking to associate clusters in different frames. 
  At each training iteration, we select two frames and apply augmentations to generate two views for each frame (i.e., $P^{1}_{m}$, $P^{1}_{n}$, $P^{2}_{m}$, $P^{2}_{n}$). A feature extractor (Backbone) is then used to obtain point-wise features in the four views, and we collect the features belonging to each cluster. In $P^{2}_{m}$, $P^{2}_{n}$, we further apply a cluster-wise pooling layer to the features to generate cluster-wise features. Finally, we minimize the distance between the point features and the corresponding cluster features from $P^{1}_{m}$, $P^{1}_{n}$, and between the cluster features obtained from associated clusters in $P^{2}_{m}$, $P^{2}_{n}$. $\tau_{1}$ and $\tau_{2}$ are data augmentations, such as random flipping and random clipping.
  }
 \vspace{-0.4cm}
  \label{fig:overview_detail}
\end{figure*}
\vspace{-0.5em}

\section{Related Work}
\vspace{-0.5em}

\label{sec:related works}


{\bf Self-supervised learning for images.} Self-supervised learning for images has developed at a fast pace in recent years~\cite{imagessl1, imagessl2 ,imagessl3, imagessl4, imagessl5, imagessl6, imagessl7}. Existing methods follow different paradigms, such as generation-based methods~\cite{generatemodel}, clustering methods~\cite{clustering1,clustering2,clustering3,clustering4} and contrastive learning methods~\cite{contrastivemode1, contrasivemodel2, contrastivemodel3}. Currently, BYOL~\cite{BYOL}, a self-supervised learning method that uses only positive pairs in its loss function, constitutes the state of the art. Intrigued by the success of such contrastive learning strategies, several works have studied the principles behind this approach, with a particular focus on the role of data augmentation~\cite{augmentation1, augmentation2, augmentation3, augmentation4}. In~\cite{augmentation5}, it was observed that data augmentation creates a certain degree of “chaos” between the intra-class samples that helps them to become more similar. Similarly, LoGo~\cite{imagessl8} also introduce local and global crops differently to handle the variance due to the augmentation.
Our method is inspired by BYOL but targets 3D data. Because of the fundamentally different nature of 2D images and 3D point clouds, data augmentation designed for images does not directly apply to the 3D domain.

{\bf Self-supervised learning for 3D data.}
As in the image case, the number of self-supervised learning methods for 3D data has grown rapidly~\cite{depthcontrast, pointcontrast, segcontrast, pointssl1, pointssl2, pointssl3, pointssl4, pointssl5}, with examples such as DepthContrast~\cite{depthcontrast}, PointContrast~\cite{pointcontrast}, GCC-3D\cite{gcc3d},  ProposalContrast~\cite{proposalcontrast}, STRL~\cite{STRL} and SegContrast~\cite{segcontrast}. Nevertheless, these methods still suffer from severe limitations. In particular, many methods~\cite{pointcontrast, STRL} need the camera pose in each frame to find correspondences to use as positive pairs. 
While effective for indoor scenes, the points in outdoor scenes are much sparser, and even with the ground-truth poses, correspondences between points are hard to obtain. By contrast, SegContrast, ProposalContrast, and DepthContrast~\cite{segcontrast,proposalcontrast, depthcontrast} specifically tackle the outdoor scenario, without requiring camera poses. However, they aggregate features in each region through either max or average pooling, and pull region-level features from different views together, therefore, it does not have constraint for each point. More importantly, they fail to find a way to associate the points in different time frames. By contrast, our method only utilizes point cloud data and does not rely on camera calibration to aggregate spatial and temporal features. Furthermore, we propose a point-to-cluster training paradigm that combines the advantages of point-level and cluster-level discrimination.

\vspace{-0.5em}

\section{Method}
\vspace{-0.5em}

\label{sec:Method}
The overall framework is depicted in Fig.~\ref{fig:overview_detail} and contains three parts: clustering and unsupervised tracking, point-wise and cluster-wise feature extraction, and spatialtemporal feature aggregation. Below, we discuss these components in detail.

\subsection{Clustering and Unsupervised Tracking}


Let $P = \left\{P^{1},P^{2},...,P^{T}\right\}$ denote a sequence of LiDAR point clouds with $T$ frames, where $P^k = \left\{p^{k}_{1},p^{k}_{2},...,p^{k}_{N_k}\right\}$ represents the $k$-th point cloud with $N_k$ 3D points $p^{k}_i \in {\mathbb{R}^{3}}$. The segmentation map of each $P^k$ is obtained by applying cluster to the non-ground points, where the ground points are eliminated by RANSAC~\cite{RANSAC}. Thanks to the over segmentation property of DBSCAN~\cite{DBSCAN}, each point has high possibility to represent the same semantic meaning with other points in the same cluster. This process yields a set of $M_k$ clusters 
$S^{k}=\left\{S_{1}^{k}, S_{2}^{k}, ..., S_{M_k}^{k} \right\}$.

We will leverage these clusters to define a point-to-cluster loss for SSL, encoding a notion of spatial similarity. 
Furthermore, we will also exploit them to create temporal positive pairs for SSL via the unsupervised tracking strategy described below. Thus, the mechanism allows similar clusters to be merged into the same one in later stage to achieve final segmentation.





Specifically, unsupervised tracking is achieved by matching the clusters in two adjacent frames, \eg, frames $k$ with $M_{k}$ clusters and $(k+1)$ with $M_{k+1}$ clusters. To this end, we define a matching degree matrix $D \in R^{M_{k} \times M_{k+1}}$ as
\begin{equation}
    D = D_{loc} + \alpha * D_{feat}, 
    \label{eq:tracking}
\end{equation}
where $D_{loc}$ is the matrix of pairwise Euclidean distances between the cluster centers in the two frames, $D_{feat}$ is the matrix of pairwise feature distance, and $\alpha \in (0, 1)$ is a weight balancing the two matrices. The center 
of cluster $j$ in frame $k$ is taken as the average of all 3D points belong to this cluster. More details regarding the cluster features is provided in Section~\ref{subsec:feature}.
We then use $D$ to match the clusters in both frames using the Hungarian algorithm\cite{hungraian}. For the unmatched clusters, we will create trajectories for the one just appears in current frame, and abandon the trajectories of the clustering no longer exists. More details can be found in the supplementary

Thanks to the combination of 3D information and learned representations, this matching strategy allows us to robustly track a cluster across multiple frames. This lets us construct long-range positive pairs where a cluster is observed under different illumination angles, thus corresponding to a challenging positive sample for SSL. We will discuss how we exploit such pairs in Section~\ref{sec:spatiotemporal}.

\vspace{-0.25em}
\subsection{Feature Extraction}
\vspace{-0.25em}

\label{subsec:feature}

As discussed above, we extract learned features from the input point clouds. Specifically, we extract two types of features: point-level ones and cluster-level ones. To this end, given an input point cloud $P^k$, we first apply data augmentation to obtain two view $\tilde{P}^{k}$ and $\bar{P}^{k}$. One view will be used to extract point-level features and the other for cluster-level features. This will let us create more challenging point-to-cluster pairs for the SSL strategy discussed in Section~\ref{sec:spatiotemporal}.


\textbf{Point-level Features.}
Following the BYOL~\cite{BYOL} format, let $f$ denote the backbone encoder. In our case, $f$ is MinkUnet~\cite{minkunet}. We forward pass $\tilde{P}^{k}$ through the backbone encoder to obtain a feature vector $y^k_{q} = f(\tilde{p}^{k}_{q})$ for every 3D point. We then group these representations according to the cluster to which each point belongs, giving us a set
$F^{k} = \left\{F^{k}_{1}, F^{k}_{2}, ..., F^{k}_{M_{k}} \right\}$, where $F^{k}_{i} \in \mathbb{R}^{N_{k,i} \times d}$, with $N_{k,i}$ the number of points in cluster $i$ from point cloud $k$, and $d$ the feature dimension of each $y^k_q$.

\textbf{Cluster-level Features.}
To extract cluster-level features, we first process $\bar{P}^{k}$ as above to extract $d$-dimensional point-level features. However, instead of simply grouping these features according to the clusters, we max-pool them according to the clusters. This yields a set of cluster-level features $C^{k} = \left\{c^{k}_1, c^{k}_{2}, ..., c^{k}_{M_{k}} \right\} $, where $c^k_i \in \mathbb{R}^{1\times d}$. 


\vspace{-0.25em}

\subsection{Spatialtemporal Feature Aggregation}
\vspace{-0.25em}

\label{sec:spatiotemporal}
Let us now describe our spatiotemporal SSL framework. It relies on two loss functions encoding two goals: \textbf {i)} points from one object should be close in feature space; \textbf {ii)} points from the same class should be closer to each other than other classes. We materialize these two objectives via the \textbf{Point-to-Cluster} and \textbf{Cluster-level Inter-frame Self-supervised Leaning} strategies discussed below.

\textbf{Point-to-Cluster Learning Strategy.} To encourage points from the same object to be close to each other, we minimize the distance between the point features of $\tilde{P}^{k}$ and the corresponding cluster features in view $\bar{P}^{k}$. Given the features discussed in Section~\ref{subsec:feature}, this is achieved via the loss function
\begin{equation}
L_{p2c} = \sum_{i=1}^{M_{k}} \sum_{j=1}^{N_{k,i}}
\left\| \frac{f^k_{i,j}}{\| f^k_{i,j} \|_2} - \frac{c^k_i}{\| c^k_i \|_2} \right\|_2^2\;,
\label{con:loss1}
\end{equation}
where $f^k_{i,j} \in \mathbb{R}^{1 \times d}$ denotes the feature vector from $F^k_{i}$ corresponding to point $j$ in cluster $i$.
In essence, this encourages the network to learn similar features for all points in the same cluster while being robust to different views of the point cloud.

\begin{table*}[t]
\centering
\begin{tabular}{ccccccccccc}
\hline
Method name       & mIoU  & car   & road   & sidewalk & building & fence & vegetation & terrain & parking & pole  \\ \hline
From scratch  & 29.17 & 82.61 & 74.32  & 52.06    & 78.99    & 19.29 & 83.13      & 68.20  & 9.04  & 30.09  \\ 
STRL \cite{STRL} &16.64 & 47.66  & 56.17 & 23.17 & 58.63    & 13.68  & 69.96 & 41.91  & 0  & 3.12  \\
DepthContrast \cite{depthcontrast} &30.91 & 88.80  & 69.51 & 49.87 & 82.67    & 22.70    & 83.36 & 67.38  & 9.32  & 48.69  \\
SegContrast  \cite{segcontrast}     & 34.01  & 89.22 & 78.72 & 57.19    & 82.80    & 21.99 & 83.42      & 67.26   & 14.06 & {\bf 50.91} \\ 
\hline
\textbf{\Methodname~(ours)} & {\bf 37.71}  & {\bf 91.11} & {\bf 85.34}  & {\bf 66.09}    & {\bf 85.43}    &  {\bf 25.63} & {\bf 84.79}  & {\bf 72.57}   & {\bf 22.61} &  48.67 \\ \hline

\end{tabular}
\caption{\bf{Per-class IoU when fine-tuning with 0.1 $\%$ labels.}}
\label{table:per class IoU}
\vspace{-0.3cm}
\end{table*}

\begin{figure*}[t]
 \centering
\includegraphics[width=1\linewidth]{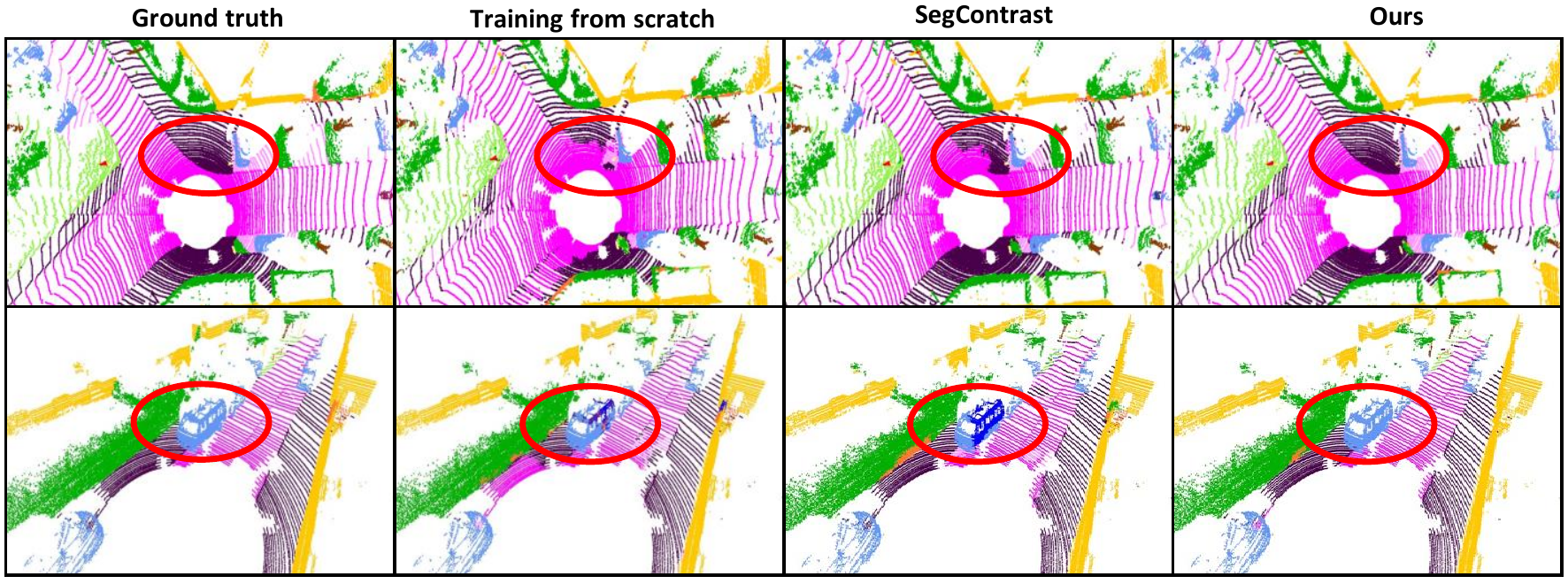}
\caption{{\bf Segmentation results on different frames (rows)}. The models are fine-tuned with 0.1\% labels on KITTI. We compare SegContrast~\cite{segcontrast}, {\Methodname~(ours)} and training from scratch (without pre-training). Our method better distinguishes the different structures shown in the highlighted area (red circle).}
\label{fig:segmentation_result}
\vspace{-0.5cm}
\end{figure*}

\textbf{Cluster-level Inter-frame Self-supervised Leaning.} To encourage points from the same class to be close to each other, we build on the observation that the main source of differences between two objects from the same class in point cloud data is the illumination angles under which they are observed. Thanks to the unsupervised tracking strategy, we can extract pairs of clusters in two distant frames, where the object is then seen under different illumination angles. Given two frames $m$ and $n$, let $N^{mn}$ denote the number of matched clusters across the two frames. Then, we use the cluster-level features to write the loss
\begin{equation}
L_{inter-frame} = \sum_{i=0}^{N^{mn}} \left\| \frac{c^m_{i}}{\| c_{i}^{m} \|_2} - \frac{c^n_{i}}{\| c^n_{i} \|_2} \right\|_2^2\;,
\end{equation}
\label{con:loss2}where $c^m_{i}$ and $c^n_{i}$ are the cluster-level feature vectors of two matched clusters in frame $m$ and $n$. Hence, the total loss can be written as
\begin{equation}\label{con:total_loss}
L_{total} = L_{p2c} + \lambda L_{inter-frame},
\end{equation}
where $\lambda$ is a weight balancing the two loss terms. In practice, the inter-frame information can be better used with the feature of SSL on intra-frame. Thus, we choose a strategy of progressively increasing the $\lambda$.

\vspace{-0.5em}
\section{Experiments}
\vspace{-0.5em}
We first describe our experimental settings, including datasets, unsupervised tracking, and implementation details. Then, we demonstrate the benefits of our self-supervised pre-trained model on downstream tasks, and finally analyze different aspects of our method.

\subsection{Experimental Settings}
\vspace{-0.25em}

\label{sec:detail}
\textbf{Datasets.} We use the KITTI~\cite{kitti} and nuScene~\cite{nuscene} datasets for pre-training, and SemanticKITTI~\cite{semantickitti} and SemanticPOSS~\cite{semanticposs} for the down-stream tasks.

KITTI \cite{kitti} has 21 sequences, and its sampling rate is 10hz. Following~\cite{segcontrast}, we use only the point clouds captured by the Velodyne LiDAR sensor rather than all the information obtained from the position sensors. The sequences 0-10 are used for pre-training, with the exception of sequence 8, which we use as validation data. 
nuScene~\cite{nuscene} is much larger than KITTI. It comprises 1000 scenes and is divided into 10 sequences. The LiDAR data is acquired at 20hz. Because of limited computational resources, we only use the point clouds captured by the Velodyne LiDAR sensor in sequence 1 and 2 (scenes 0 - 149) for pre-training. 
SemanticKITTI~\cite{semantickitti} provides dense point-wise annotations for almost every point in KITTI~\cite{semantickitti}. A total of 23,201 scans are annotated on sequences 0-10 of KITTI for training and validation.
SemanticPOSS \cite{semanticposs} is also used for semantic segmentation, and contains 2988 diverse and complicated LiDAR scans with 14 classes. The scans are divided into 6 splits, with 500 scans per split. Splits 4 and 5 are used for testing and the other ones are used for training. 

{\bf Unsupervised Tracking.} We set the relative weight between spatial distance and feature distance in~\cref{eq:tracking} to $\alpha = 0.5$, and the threshold of RANSAC distance to 0.25 as in \cite{segcontrast}. The threshold of DBSCAN distance is set to 0.25 in KITTI and 0.5 in nuScene. In each frame, we drop clusters with fewer than 200 points or more than 20000 points to filter out noise and retain up to 50 clusters.

\textbf{Implementation Details.} We compare our approach with DepthContrast~\cite{depthcontrast}, STRL~\cite{STRL}, SegContrast~\cite{segcontrast}, and training from scratch. We use MinkUnet~\cite{minkunet} as backbone for all approaches and build our approach on the basis of BYOL\cite{BYOL}. 
We pre-train the backbone on KITTI and nuScene for 200 epochs using an SGD optimizer with a momentum of 0.9, and set the weight decay to 0.0004 following SegContrast~\cite{segcontrast}. The learning rate is initially set to 0.036 with a linear annealing scheme with a minimum learning rate equal to 0.009. In the early training stages, we set $\lambda$ to 0, and to 4 in the later stages. When incorporating the inter-frame loss term, we re-initialize different MLPs after the backbone networks to avoid information leaks from spatial. The batch size is set to 8 for each GPU, and we use $8 \times$GTX3090 GPUs to pre-train the models, which leads to the total batch size of 64.

For fine-tuning on the down-stream semantic segmentation task, we use an SGD optimizer with a cosine learning rate schedule. The fine-tuned models are evaluated on the validation sequences, i.e., sequence 8 for SemanticKITTI and sequences 4 and 5 for SemanticPOSS~\cite{segcontrast}. The batch size is set to 2 for each GPU, and $4 \times$GTX2080Ti GPUs are used for the experiments.

\textbf{Evaluation Metrics.} We evaluate point cloud semantic segmentation using the mean intersection over
union (mIoU) and the overall point classification accuracy (Acc).


\begin{table}[h]

\centering
\begin{tabular}{ccccc}
\hline
                          & 0.1\% & 1\%    & 10\%         & 100\% \\ \hline
From Scratch                & 29.17 & 48.11  & 51.00        & 56.14 \\
STRL\cite{STRL}                  & 16.64 & 31.88      & 30.88   & 55.71 \\
DepthContrast \cite{depthcontrast}              & 30.91 & 42.41  & 42.38        & 45.48 \\
SegContrast \cite{segcontrast}                & 34.01 & 48.02  & 52.26        & 55.45 \\ \hline
\textbf{\Methodname~(ours) }                      & {\bf 37.71}    & {\bf52.60}   & {\bf54.51}       & {\bf57.33} \\ \hline
\end{tabular}
\vspace{-0.2cm}

\caption{Pre-training on {\bf KITTI} and evaluating the fine-tuned models in different label regimes on \textbf{SemanticKITTI} for semantic segmentation. We report the mIoU.}
\label{table:finetune_at_kitti_segmentation}
\vspace{-0.2cm}

\end{table}

\begin{table}[h]
\centering
\tabcolsep=0.35cm
\begin{tabular}{ccc}
\hline
mIoU / Acc               & seq1              & seq 1-2 \\ \hline
From Scratch       & 29.17 / 82.57                     & 29.17 / 82.57 \\
STRL\cite{STRL}                    & 19.11 / 74.56         & 18.74 / 70.85    \\
SegContrast \cite{segcontrast}               & 33.91 / 84.88         & 34.28 / 85.20    \\ \hline
\textbf{\Methodname~(ours) }         & {\bf 34.43 / 85.34 }        & {\bf 35.08 / 85.75} \\ \hline
\end{tabular}
\vspace{-0.2cm}

\caption{Pre-training on \textbf{nuScene} and evaluating the fine-tuned models in the 0.1\% label regime on \textbf{SemanticKITTI} for semantic segmentation. We report mIoU/Acc.}
\label{table:nuscene_finetune_at_kitti_segmentation}
\vspace{-0.2cm}
\end{table}



\begin{table}[h]
\centering
\tabcolsep=0.3cm

\begin{tabular}{ccc}

\hline pre-train dataset & KITTI      & nuScene(seq 1)  \\ \hline
From Scratch                & 39.64 / 88.66      & 39.64 / 88.66 \\
STRL\cite{STRL}                     & 38.43 / 88.32      & 36.63 / 87.53 \\ 
SegContrast \cite{segcontrast}      & {\bf 43.88 /  89.64}      & 42.86 / 89.28 \\ \hline
\textbf{\Methodname~(ours) }     &  43.84 / 89.47 & {\bf 43.55 / 89.38}  \\ \hline
\end{tabular}
\vspace{-0.2cm}

\caption{Pre-training on \textbf{KITTI and nuScene}, and evaluating the fine-tuned models on \textbf{SemanticPOSS} for semantic segmentation. We report the mIoU/Acc.}
\label{table:kittt and nuscene finetune at poss segmentation}
\vspace{-0.2cm}

\end{table}


\begin{table}[]
\centering
\tabcolsep=0.3cm
\begin{tabular}{ccc}
\hline
mIoU / Acc                           & 0.1\%  & 1\%    \\ \hline
From Scratch                & 29.17 / 82.57 & 48.11 / 89.94\\
STRL\cite{STRL}                  & 16.64 / 69.36 & 31.88 / 85.35      \\
DepthContrast \cite{depthcontrast}              & 30.91 / 82.82 & 42.41 / 88.95  \\
SegContrast \cite{segcontrast}                & 34.01 / 84.72 & 48.02 / 88.84  \\ \hline
P2C     & {\bf 35.48 / 86.18}& {\bf 50.83 / 90.14} \\ \hline
\end{tabular}
\vspace{-0.2cm}

\caption{Ablation study on the \textbf{pre-training strategy} with 0.1\% and 1\% labels. We report mIoU/Acc.}
\label{table: ablation on frameworks}
\vspace{-0.6cm}
\end{table}

\vspace{-0.25em}
\subsection{Outdoor Scene Understanding}
\vspace{-0.25em}

\label{sec:Outdoor_scene}


{\bf  Label Efficiency.} To assess the label efficiency of our \Methodname~ approach, we fine-tune the model pre-trained on KITTI on SemanticKITTI. Following~\cite{segcontrast}, SemanticKITTI is divided into different regimes corresponding to different percentages of labels. Specifically, we use 0.1\%, 1\%, 10\%, and 100\% of the training data to fine-tune the pre-trained model for semantic segmentation.

In Table \ref{table:per class IoU}, we compare the mIoU and per-class IoU of the proposed \Methodname~ and of the state-of-the-art approaches when using 0.1\% of the labels. Our method outperforms the baselines by an impressive margin, yielding an mIoU of 37.71\%, which is 3.7\% better than SegContrast\cite{segcontrast} and 8.54\% than training from scratch. Fig.~\ref{fig:segmentation_result} evidences that our method yields more complete and accurate segmentation masks than the other methods.


The per-class comparison shows that our approach greatly improves the network's performance on most classes when there are few annotations. Our {\Methodname} yields much better results than the baselines, especially for car, building, vegetation, terrain, and parking.
We attribute this to the fact that these classes have clearly different appearances under different illumination angles as shown in Fig.\ref{fig:angle_different_1}, which is exactly the problem that our inter-frame self-supervised learning addresses.


The results obtained by fine-tuning with different percentages of training data are provided in Table~\ref{table:finetune_at_kitti_segmentation}. Our method consistently outperforms the state-of-the-art self-supervised approach for all label regimes. Specifically, the mIoU of our approach outperforms the SegContrast and From Scratch ones by 2.25\% and 3.51\%, with 10\% labels.

  

\begin{figure*}[t]
 \centering
\includegraphics[width=1\linewidth]{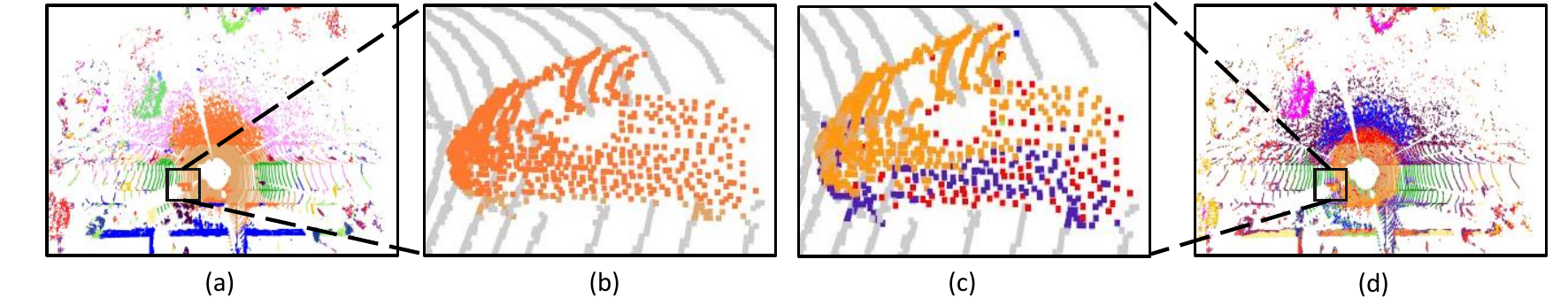}
\vspace{-0.6cm}
\caption{\textbf{Comparison of the features generated by different pre-trained models}. (a) Results with P2C. (b) Zoomed in car from (a). (c) Zoomed in car from (d). (d) Results with SegContrast\cite{segcontrast}. The points with the same color are in the same cluster(clustering in feature space). The colors of (a, b) and (c, d) are independent and have no relationship with each other. For better visualization, we only colorize the car in (b) and (c). }
\label{fig:p2s_feature_map}
\vspace{-0.3cm}
\end{figure*}

\begin{table*}[t]
\centering
\begin{tabular}{ccccccccccc}
\hline
Method name       & mIoU  & car   & road   & sidewalk & building & fence & vegetation & terrain & parking & pole  \\ \hline
SegContrast  \cite{segcontrast}     & 34.01  & 89.22 & 78.72 & 57.19    & 82.80    & 21.99 & 83.42      & 67.26   & 14.06 & 50.91 \\ 
SegContrast-BYOL & 33.94  & 89.34 & 78.56 & 57.29    & 82.86 & 21.19 & 83.26      & 67.30   & 14.09 & 50.54 \\ 
SegContrast-Inter &  36.03  &  90.52 &  81.45  &  62.90  &  83.54    &  21.46 &  84.37    & 72.52   & 17.05 &  {\bf 51.78} \\ 
P2C  & 35.48  & 90.18 & 80.90 & 61.91    & 83.92    & 25.45 & 84.30  & 71.35  & 18.73 & 46.95 \\\hline
\textbf{\Methodname~(ours)} & {\bf 37.71}  & {\bf 91.11} & {\bf 85.34}  & {\bf 66.09}    & {\bf 85.43}    &  {\bf 25.63} & {\bf 84.79}  & {\bf 72.57}   & {\bf 22.61} &  48.67 \\ \hline
\end{tabular}
\vspace{-0.2cm}
\caption{\textbf{Per-class IoU when fine-tuning with 0.1 $\%$ labels}. SegContrast-BYOL: SegContrast built on the basis of BYOL. SegContrast-Inter: SegContrast with a additional inter-frame self-supervised learning stage after the original SegContrast. P2C: Our approach with only the point-to-cluster loss function.}
\label{table:per_class_IoU2}
\vspace{-0.6cm}
\end{table*}

{\bf Feature Representation Transferability.} To confirm the transferability of the features learned by our approach, 
we pre-train our models on nuScene and design two settings for pre-training: i) only using sequence 1 (seq1), and ii) using sequence 1 and 2 (seq 1-2) with uniform down-sampling to keep the number of frames consistent with seq1 and the frame rate consistent with KITTI.

As shown in Table~\ref{table:nuscene_finetune_at_kitti_segmentation}, our method outperforms training from scratch and Segcontrast~\cite{segcontrast} when using only seq 1 from nuScene for pre-training and fine-tuning on SemanticKITTI~\cite{semantickitti}. Our approach improves the segmentation performance by 5.26\% in mIoU and 2.77\% in Acc. When seq 1 and 2 are used for pre-training, our approach improves the segmentation performance by 5.91\% in mIoU and 3.18\% in Acc. We also fine-tune the models pre-trained on nuScene or KITTI to the semantic segmentation task using SemanticPOSS~\cite{semanticposs}. 
As SemanticPOSS is small, we fine-tune on the entire dataset. Table~\ref{table:kittt and nuscene finetune at poss segmentation} shows that our method yields better mIoU results than training from scratch. When the network is pre-trained on KITTI, our approach improves the mIoU by 4.20\% compared to the network without pre-training.

\begin{figure*}[t]
  \centering
  \includegraphics[width=1\linewidth]{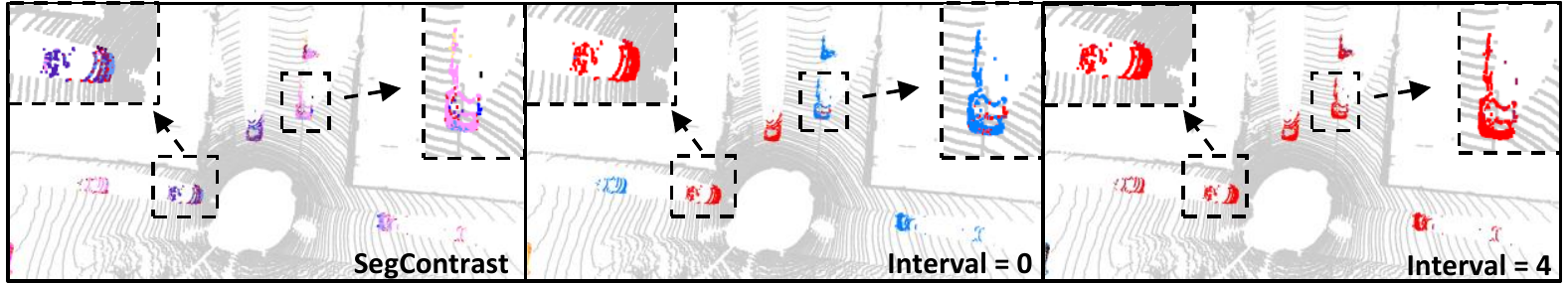}
  \caption{\textbf{Comparison of models pre-trained with different intervals between two frames}. Left: SegContrast. Middle: (\Methodname, interval=0). Right: (\Methodname, interval=4). The pre-trained models are used to extract point features for visualization, and the points with the same color are adjacent in feature space. For better visualization, we only colorize the points belonging to the specified clusters. }
  \vspace*{-0.5cm}
  \label{fig:track_feature_show}
\end{figure*}

\vspace{-0.25em}
\subsection{Analysis}
\vspace{-0.25em}
\label{sec:analysis}
\textbf{Guaranteed over-segmentation assumptions.}
 P2C SSL strategy relies on the over-segmentation assumptions and 
the hyper-parameter leading to the over-segmented clusters is easy to set. To show this, we performed the following experiment. We varied the DBSCAN distance threshold from 0.15 to 0.45 and measured the proportion of clusters having at least 90 $\%$ of their points from the same semantic class. The higher this proportion, the higher the chance that the clusters are over-segmented and not under-segmented. In the worst case, we found 73.45$\%$ of the clusters being over-segmented witch evidences the stability of the hyper-parameter of DBSCAN. 

\begin{figure}[t]
  \centering
  \includegraphics[width=1\linewidth]{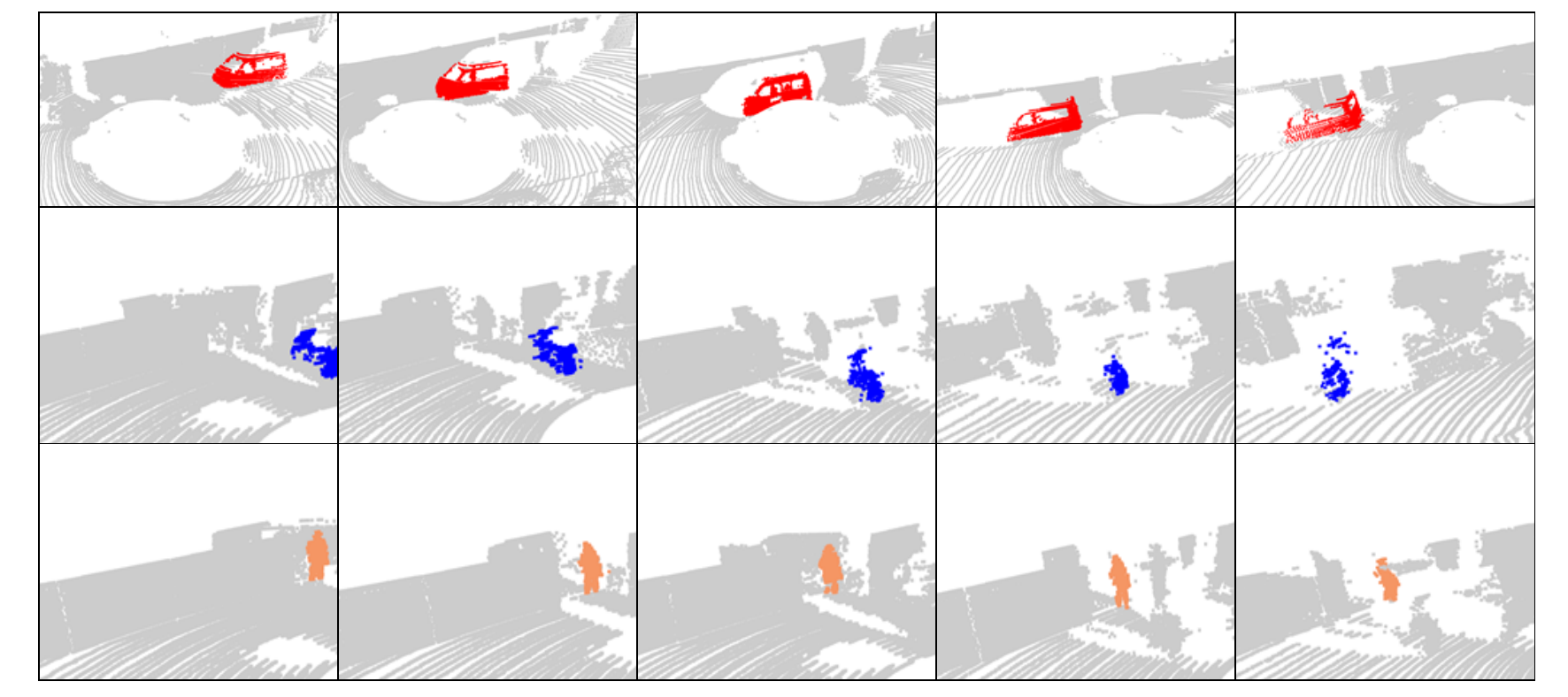}
    \vspace*{-1em}
  \caption{{\bf Associated clusters in multiple frames}. Some clusters with clear semantic information are selected for display. The same color represents the same cluster. Red: car; Blue: motorcycle; Orange: person.}
  \vspace*{-1em}
  \label{fig:track_show}
\end{figure}

\textbf{Performance of unsupervised tracking.}
We measured that 63.73$\%$ of the clusters are tracked for at least 3 frames, and 31.33$\%$ for at least 8 frames. Such tracking times are sufficient for us to create positive pairs of clusters observed under different illumination angles as shown in Fig.~\ref{fig:track_show}.

\vspace*{-0.5em}
\subsection{Ablation Study}
  \vspace*{-0.5em}

\label{sec:ablation}

\begin{figure}[h]
  \centering
  \includegraphics[width=1\linewidth]{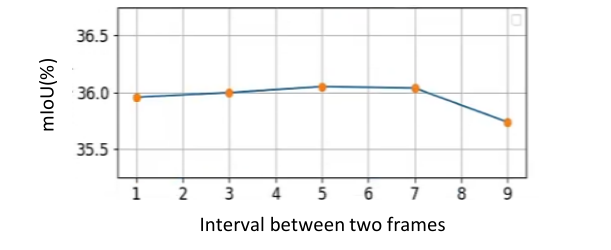}
  \vspace{-0.8cm}
  \caption{Comparison mIoU on SemanticKITTI dataset, which is fine-tuned by the models pre-trained with \textbf{different intervals} between two frames.}
  \vspace{-0.3cm}
  \label{fig:different_frames}
\end{figure}

Experiment in this section, if it is not specially stated, the model pre-training on KITTI, and fine-tuning and evaluating on SemanticKITTI.

\textbf{Scene- vs. Cluster-level Pre-training.}
To demonstrate the effectiveness of the Point-to-Cluster learning strategy proposed in Section~\ref{sec:spatiotemporal}, we conduct an experiment with only the point-to-cluster loss function of
~\cref{con:loss1} for pre-training. We dub this setting P2C. As shown in Table~\ref{table: ablation on frameworks} and Table~\ref{table:per_class_IoU2}, a cluster-level pre-trained SegContrast performs better than scene-level pre-trained DepthContrast and STRL. Our proposed P2C pre-training method achieves better results than SegContrast. To better understand the advantages of P2C, we have designed the following visualization. We select a point cloud frame, use the pre-trained model (i.e., SegContrast and P2C) to extract features for each point, and use K-Means~\cite{Kmeans} to cluster these features. Specifically, we use 20 clusters, which corresponds to the number of categories in the annotations. 
In Fig.~\ref{fig:p2s_feature_map}, we visualize the points by coloring them according to the clusters. 
With features extracted by the SegContrast pre-trained model, the car zoomed in in Fig.~\ref{fig:p2s_feature_map} (c) is divided into several colors. This indicates that the points from the same category can be distant in the learned feature space.
By contrast, encouraging the points in the same cluster to have similar features, 
P2C yields feature such that most of the car points are in the same cluster, as shown in Fig.~\ref{fig:p2s_feature_map} (b), thus indicating that the features are more representative of the object instances.




\vspace{-0.2em}

\textbf{Effectiveness of Inter-frame Self-supervised Learning.}
To better demonstrate the role of inter-frame self-supervised learning, we add a new stage after the original SegContrast, in which we activate our inter-frame loss function, as in our method. Note that we maintain the total number of training epochs to 200.
We dub the resulting model SegContrast-inter, and show its per-class IoU in Table~\ref{table:per_class_IoU2}. For most classes, SegContrast-inter performs much better than SegContrast. However, SegContrast performs better in the fence class. Fence has almost the same appearance  when they are viewed from different angles (the z axis remains unchanged). This experiment also supports our motivation that the illumination angle is an important factor, making the appearance of an object differ. 

\textbf{Interval between Two Frames.} 
Since we choose positive pairs from different frames, the interval between the two selected frames is a hyper-parameter affecting the results. In our approach, we gradually increase the interval between two frames to avoid having to set such a hyper-parameter. Here we evaluate the impact of the interval on a model using a fixed interval between two frames. The experiments are repeated for 3 times to reduce randomness, and we report the average. As shown in Fig.~\ref{fig:different_frames}, 
with the increase of interval, the performance first increases and then decreases. We think it is because greater interval can reduce the impact of more illumination angles, but the number of clusters that can be tracked across frames decreases as the interval increases.  The best-performing model corresponds to an interval of 5, reaching an mIoU of 36.05\%  when it reaches the balance of illumination angles change and number of tracking clusters. Note that the worst performance (interval of 9) is still better than that of SegContrast. 
 
In Fig.~\ref{fig:track_feature_show}, we visualize a frame with multiple cars under different LiDAR illumination angles.
We also render the points in the same cluster with the same color. 
Note that SegContrast and {\Methodname} with interval=0 cannot extract similar features for the cars under different illumination angles. By contrast, {\Methodname} with interval=4 yields similar features for all cars, benefiting from inter-frame matching.

\begin{table}[]
\centering
\tabcolsep=0.35cm

\begin{tabular}{ccc}
\hline
mIoU / Acc              & 0.1 \%      & 1 \% \\ \hline 
From Scratch         & 29.17 / 82.57     & 48.11 / 89.94 \\
SegContrast \cite{segcontrast}                & 34.01 / 84.72 & 48.02 / 88.84  \\
{\textbf{\Methodname-n} }    & 36.01 / 87.44     & 51.06 / 89.68 \\  \hline
{\textbf{\Methodname}}       & {\bf 37.71 / 87.67} & {\bf 52.60 / 90.24}  \\ \hline
\end{tabular}
\caption{\textbf{Ablation study on feature similarity for tracking}. {\Methodname-n} only considers location similarity in tracking stage.}
\label{table: ablation on match_byol}
\vspace{-2 em}

\end{table}

\textbf{Tracking with vs. without a Model.}
In ~\cref{eq:tracking}, we use both location and feature similarities to compute the matching degree matrix for tracking. To illustrate the effectiveness of the feature similarity, we re-conduct the tracking with only location similarity. This is indicated by {\Methodname-n} in Table \ref{table: ablation on match_byol}.
Note that the resulting method still outperforms SegContrast but not our complete \Methodname. 

\begin{table}[t]
\centering
\tabcolsep=0.46cm

\begin{tabular}{ccc}
\hline
Method           & mIoU (\%)       & Acc (\%)   \\ \hline
SegContrast\cite{segcontrast}      & 34.01          & 84.72          \\
SegContrast-BYOL & 33.94          & 84.67          \\ \hline
\textbf{\Methodname~(Ours)} & {\bf 37.71}    & {\bf 87.67} \\ \hline
\end{tabular}
\caption{\textbf{Ablation study on the framework}. SegContrast-BYOL: SegContrast built on the basis of BYOL.}
\label{table: ablation on framework}
\vspace{-0.6cm}

\end{table}

\textbf{Effect of BYOL.}
To evidence that the benefits of our approach over SegContrast are not only due to our use of BYOL instead of MoCo but truly to our training formalism, we replace the MoCo in SegContrast with BYOL. As shown in Table~\ref{table: ablation on framework}, SegContrast-BYOL performs on par with the original SegContrast (with MoCo) in the 0.1\% label regime. It implies that the networks are not the key factor in point clouds SSL. Due to the over-segmentation, we used in our method, where each cluster could belong to the same class, building negative pairs for each point in our point-to-cluster strategy will harm the optimization. By contrast, it is more intuitive for us to only build positive pairs to avoid pushing the nearby clusters away, which offers a chance to merge the cluster with the help of temporal information.



\vspace{-0.5em}
\section{Conclusion and Limitation}
\vspace{-0.5em}


In this paper, we have introduced an SSL strategy for point cloud segmentation without external supervision. It relies on a novel Point-to-Cluster (P2C) training paradigm to exploit spatial information, and further introduces an inter-frame self-supervised learning strategy to capture temporal information. Altogether, our approach provides a practical tool for pre-training with point clouds in the wild. Experiments evidence that our approach outperforms the state-of-the-art SSL techniques for point cloud in the wild. However, such an improvement does not materialize for objects whose appearance is invariant to viewpoint changes, such as fences. In the future, we will therefore focus on how to improve the segmentation ability of such objects.

\small{\textbf{Acknowledgement.} This work was supported in part by the National Natural Science Foundation of China under Grant No. 62006182 and the Swiss National Science Foundation via the Sinergia grant CRSII5-180359.}

{\small
\bibliographystyle{ieee_fullname}
\bibliography{egbib}
}

\end{document}